\documentclass[11pt,a4paper]{article}
\usepackage[hyperref]{emnlp-ijcnlp-2019}
\usepackage{times}
\usepackage{latexsym}
\usepackage{url}
\usepackage{listings}
\usepackage[utf8]{inputenc}
\usepackage[vietnamese,english]{babel}
\usepackage{CJKutf8}
\usepackage[T1]{fontenc}

\usepackage{multirow}
\usepackage{color}
\usepackage{listings}
\usepackage{amsmath}

\definecolor{codegreen}{rgb}{0,0.6,0}
\definecolor{codegray}{rgb}{0.5,0.5,0.5}
\definecolor{codepurple}{rgb}{0.58,0,0.82}
\definecolor{backcolour}{rgb}{0.95,0.95,0.92}

\lstdefinestyle{mystyle}{
    backgroundcolor=\color{backcolour},   
    commentstyle=\color{codegreen},
    keywordstyle=\color{magenta},
    numberstyle=\tiny\color{codegray},
    stringstyle=\color{codepurple},
    basicstyle=\footnotesize,
    breakatwhitespace=false,         
    breaklines=true,                 
    captionpos=b,                    
    keepspaces=true,                            
    showspaces=false,                
    showstringspaces=false,
    showtabs=false,                  
    tabsize=2
}
\aclfinalcopy 

\newcommand{\vect}[1]{\mathbf{#1}}

\title{Overcoming the Rare Word Problem for Low-Resource Language Pairs \\ in Neural Machine Translation}

\author{
\begin{tabular}{c} Thi-Vinh Ngo \\
  Thai Nguyen University \\ 
  {\tt ntvinh@ictu.edu.vn} \\ \\
  Phuong-Thai Nguyen \\
  Vietnam National University \\
  {\tt thainp@vnu.edu.vn} \end{tabular} \\ \And
 \begin{tabular}{c} Thanh-Le Ha \\
  Karlsruhe Institute of Technology \\
  {\tt thanh-le.ha@kit.edu}  \\ \\ 
  Le-Minh Nguyen \\
  JAIST, Japan \\
  {\tt nguyenml@jaist.ac.jp} \end{tabular} 
  }

\date{}

\begin{document}
\maketitle
\begin{abstract}
Among the six challenges of neural machine translation (NMT) coined by \cite{Koehn2017}, rare-word problem is considered the most severe one, especially in translation of low-resource languages. In this paper, we propose three solutions to address the rare words in  neural machine translation  systems. First, we enhance source context to predict the target words by connecting directly the source embeddings to the output of the attention component in NMT. Second, we propose an algorithm to learn morphology of unknown words for English in supervised way in order to minimize the adverse effect of rare-word problem. Finally, we exploit synonymous relation from the WordNet to overcome out-of-vocabulary (OOV) problem of NMT. We evaluate our approaches on two low-resource language pairs: English-Vietnamese and Japanese-Vietnamese. In our experiments, we have achieved significant improvements of up to roughly +1.0 BLEU points in both language pairs.

\end{abstract}

\section{Introduction}
NMT systems have achieved better performance compared to statistical  machine translation (SMT) systems in recent years not only on available data language pairs~\cite{sennrich2016edinburgh,cho2016adaptation},  but also on low-resource language pairs~\cite{Toan2017,cettolo2016iwslt}. Nevertheless, NMT still exists many challenges which have adverse effects on its effectiveness \cite{Koehn2017}. One of these challenges is that NMT has biased tend in translating high-frequency words, thus words which have lower frequencies are often translated incorrectly. This challenge has also been confirmed again in~\cite{Toan2017}, and they have proposed two strategies to tackle this problem with modifications on the model's output distribution: one for normalizing some matrices by fixing them to constants after several training epochs and another for adding a direct connection from source embeddings through a simple feed forward neural network (FFNN). These approaches increase the size and the training time of their NMT systems. In this work, we follow their second approach but simplify the computations by replacing FFNN with two single operations.

Despite above approaches can improve the prediction of rare words, however, NMT systems often use limited vocabularies in their sizes, from 30K to 80K most frequent words of the training data, in order to reduce computational complexity and the sizes of the models~\cite{Bahdanau2014,Luong2015b}, so the rare-word translation are still problematic in NMT. Even when we use a larger vocabulary, this situation still exists~\cite{Jean2015}. A word which has not seen in the vocabulary of the input text (called~\textit{unknown word}) are presented by the $unk$ symbol in NMT systems. Inspired by alignments and phrase tables in phrase-based machine translation (SMT) as suggested by \cite{Koehn2007}, \cite{Luong2015b} proposed to address OOV words using an annotated training corpus. They then used a dictionary generated from alignment model or maps between source and target words to determine the translations of $unks$ if translations are not found. \cite{Sennrich2016} proposed to reduce unknown words using Gage's Byte Pair Encoding (BPE) algorithm \cite{Gage1994}, but NMT systems are less effective for low-resource language pairs due to the lack of data and also for other languages that sub-word are not the optimal translation unit. In this paper, we employ several techniques inspired by the works from NMT and the traditional SMT mentioned above. Instead of a loosely unsupervised approach, we suggest a supervised approach to solve this trouble using synonymous relation of word pairs from WordNet on Japanese$\rightarrow$Vietnamese and English$\rightarrow$Vietnamese systems. To leverage effectiveness of this relation in English, we transform variants of words in the source texts to their original forms by separating their affixes collected by hand. 

Our contributes in this work are:
\vspace*{-0.2cm}
\begin{itemize}
\setlength{\itemsep}{0pt}
  \item We release the state-of-the-art for Japanese-Vietnamese NMT systems.
  \item We proposed the approach to deal with the rare word translation by integrating source embeddings to the attention component of NMT.
  \item We  present a supervised algorithm to reduce the number of unknown words for the  English$\rightarrow$Vietnamese translation system.
  \item We demonstrate the effectiveness of leveraging linguistic information from WordNet to alleviate the rare-word problem in NMT.
\end{itemize}


\section{Neural Machine Translation}
Our NMT system use a bidirectional recurrent neural network (biRNN) as an encoder and a single-directional RNN as a decoder with input feeding of \cite{Luong2015a} and the attention  mechanism of \cite{Bahdanau2014}. The Encoder's biRNN are constructed by two RNNs with the hidden units in the LSTM cell, one for forward and the other for backward of the source sentence $\vect{x}=(x_1, ...,x_n)$. Every word $x_i$ in sentence  is first encoded into a continuous representation $E_s(x_i)$, called the source embedding. Then $\vect{x}$ is transformed into a fixed-length hidden vector $\vect{h}_i$ representing the sentence at the time step $i$, which called the annotation vector, combined by the states of forward $\overrightarrow{\vect{h}}_i$ and backward $\overleftarrow{\vect{h}}_i$: 
\begin{center}
   $\overrightarrow{\vect{h}}_i=f(E_s(x_i),\overrightarrow{\vect{h}}_{i-1})$\\
   $\overleftarrow{\vect{h}}_i=f(E_s(x_i),\overleftarrow{\vect{h}}_{i+1})$
\end{center} 

The decoder generates the target sentence $\vect{y}={(y_1, ..., y_m)}$, and at the time step $j$, the predicted probability of the target word $y_j$ is estimated as follows:
$$
p(y_j|y_{<j}, \vect{x}) \propto \text{softmax}(\vect{W}\vect{z}_j + \vect{b}) 
$$
where $\vect{z}_j$ is the output hidden states of the attention mechanism and computed by the previous output hidden states $\vect{z}_{j-1}$, the embedding of previous target word $E_t(y_{j-1})$ and the context $\vect{c}_j$:
\begin{center}
  $\vect{z}_j=g(E_t(y_{j-1}), \vect{z}_{j-1}, \vect{c}_j)$ 
\end{center}

The source context $\vect{c}_j$ is the weighted sum of the encoder's annotation vectors $\vect{h}_i$:
\begin{center}
  $\vect{c}_j=\sum^n_{i=1}\alpha_{ij}\vect{h}_i$
\end{center}
where $\alpha_{ij}$ are the alignment weights, denoting the relevance between the current target word $y_j$ and all source annotation vectors $\vect{h}_i$.
\section{Rare Word translation}
\label{rare-word-translation}
In this section, we present the details about our approaches to overcome the rare word situation. While the first strategy augments the source context to translate low-frequency words, the remaining strategies reduce the number of OOV words in the vocabulary.

\subsection{Low-frequency Word Translation}
\label{source-emb}
The attention mechanism in RNN-based NMT maps the target word into source context corresponding through the annotation vectors $\vect{h}_i$. In the recurrent hidden unit, $\vect{h}_i$ is computed from the previous state $\vect{h}_{t-1}$. Therefore, the information flow of the words in the source sentence may be diminished over time. This leads to the accuracy reduction when translating low-frequency words, since there is no direct connection between the target word and the source word. To alleviate the adverse impact of this problem, \cite{Toan2017} combined the source embeddings with the predictive distribution over the output target word in several following steps:

Firstly, the weighted average vector of the source embeddings is computed as follows:
$$
\vect{l}_t = \tanh{\sum_e \alpha_j(e) \vect{f}_e} 
$$
where $\alpha_j(e)$ are alignment weights in the attention component and $f_e = E_s(x)$, are the embeddings of the source words.

Then $l_j$ is transformed through one-hidden-layer FFNN with residual connection proposed by \cite{He2015}:
$$
\vect{t}_j = \tanh({\vect{W}_l\vect{l}_j}) + \vect{l}_j 
$$
Finally, the output distribution over the target word is calculated by:
$$
p(y_j|y_{<j}, \vect{x}) = \text{softmax}(\vect{W}\vect{z}_j + \vect{b} + \vect{W}_t\vect{t}_j + \vect{b}_t) 
$$
The matrices $\vect{W}_l$, $\vect{W}_t$ and $\vect{b}_t$ are trained together with other parameters of the NMT model.

This approach improves the performance of the NMT systems but introduces more computations as the model size increase due to the additional parameters $\vect{W}_l$, $\vect{W}_t$ and $\vect{b}_t$. We simplify this method by using the weighted average of source embeddings directly in the softmax output layer:
$$
p(y_j|y_{<j}, \vect{x}) = \text{softmax}(\vect{W}(\vect{z}_j + \vect{l}_j)+ \vect{b})
$$
Our method does not learn any additional parameters. Instead, it requires the source embedding size to be compatible with the decoder's hidden states. With the additional information provided from the source embeddings, we achieve similar improvements compared to the more expensive method described in \cite{Toan2017}.
\subsection{Reducing Unknown Words}
In our previous experiments for English$\rightarrow$Vietnamese, BPE algorithm~\cite{Sennrich2016} applied to the source side does not significantly improves the systems despite it is able to reduce the number of unknown English words. We speculate that it might be due to the morphological differences between the source and the target languages (English and Vietnamese in this case). The unsupervised way of BPE while learning sub-words in English thus might be not explicit enough to provide the morphological information to the Vietnamese side. In this work, we would like to attempt a more explicit, supervised way. We collect 52 popular affixes (prefixes and suffixes) in English and then apply the separating affixes algorithm (called \textbf{\textit{SAA}}) to reduce the number of unknown words as well as to force our NMT systems to learn better morphological mappings between two languages. 

The main ideal of our \textbf{\textit{SAA}} is to separate affixes of unknown words while ensuring that the rest of them still exists in the vocabulary. Let the vocabulary $V$ containing $K$ most frequency  words from the training set $T1$, a set of prefixes $P$, a set of suffixes $S$, we call word $w'$ is the rest of an unknown word or rare word $w$ after delimiting its affixes. We iteratively pick a $w$ from $N$ words (including unknown words and rare words) of the source text $T2$ to consider if $w$ starts with a prefix $p$ in $P$ or ends with a suffix $s$ in $S$, we then determine splitting its affixes if $w'$ in $V$. A rare word in $V$ also can be separated its affixes if its frequency is less than the given threshold. We set this threshold by $2$ in our experiments. Similarly to BPE approach, we also employ a pair of the special symbol $@$ for separating affixes from the word. Listing~\ref{SAA} shows our \textit{\textbf{SAA}} algorithm.

\noindent\begin{minipage}{\linewidth}
\begin{lstlisting}[language=Python, title=The proposed SAA for separating affixes from words.]
Input: T1, T2, P, S, threshold=1
Output: the output text T

V = get_most_frequency_K_words(T1)
N = get_words_from_the_source_text(T2)
T = T2

for each word w in N:
 if w not in V or freq(w) <= threshold:
   for each prefix p in P:
     w1 = separate_prefix(p)
     if w1 != w and w1 in V:
       T = replace(T, w, w1, p)
       break
   for each suffix s in S: 
     w2 = separate_suffix(s)
     if w2 != w1 and w2 in V:
       T = replace(T, w2, w1, s)
       break
return  T

Example: intercepted -> intercept @@ed
         impulsively -> impulsive @@ly
         overlooks -> over@@ look @@s
         disowned -> dis@@ own @@ed
\end{lstlisting}
\label{SAA}
\end{minipage}
\subsection{Dealing with OOV using WordNet}
WordNet is a lexical database grouping words into sets which share some semantic relations. Its version for English is proposed for the first time by \cite{Fellbaum1998}. It becomes a useful resource for many tasks of natural language processing \cite{Kolte2008,Mendez2013,Gao2014}. WordNet are available mainly for English and German, the version for other languages are being developed including some Asian languages in such as Japanese, Chinese, Indonesian and Vietnamese. Several works have employed WordNet in SMT systems\cite{Khodak2017, Arcan2019} but to our knowledge, none of the work exploits the benefits of WordNet in order to ease the rare word problem in NMT. In  this work, we propose the learning synonymous algorithm (called \textbf{\textit{LSW}}) from the WordNet of English and Japanese to handle unknown words in our NMT systems.

In WordNet, synonymous words are organized in groups which are called synsets. Our aim is to replace an OOV word  by its synonym which appears in the vocabulary of the translation system. From the training set of the source language $T1$, we extract the vocabulary $V$ in size of $K$ most frequent words. For each OOV word from  $T1$,  we learn its synonyms which exist in the $V$ from the WordNet $W$. The synonyms are then arranged in the descending order of their frequencies to facilitate selection of the $n$ best words which have the highest frequencies. The output file $C$ of the algorithm contains OOV words and its corresponding synonyms and then it is applied to the input text $T2$. We also utilize a frequency threshold for rare words in the same way as in \textit{\textbf{SAA}} algorithm. In practice, we set this threshold as $0$, meaning no words on $V$ is replaced by its synonym. If a source sentence has $m$ unknown words and each of them has $n$ best synonyms, it would generate $m^n$ sentences. Translation process allow us to select the best hypothesis based on their scores.  Because of each word in the WordNet can belong to many synsets with different meanings, thus an inappropriate word can be placed in the current source context. We will solve this situation in the further works. Our systems only use 1-best synonym for each OOV word.  Listing~\ref{LSW} presents the \textit{\textbf{LSW}} algorithm.
\noindent\begin{minipage}{\linewidth}
\begin{lstlisting} [mathescape=true, language=Python, title=The LSW learns synonymous words from WordNet.]
Input: T1, T2, $W_s$, threshold=1
Output: - C: The list contains synonymous words for OOV words.
        - T: The input of the translation systems
        
def learn_synonym()
 V=get_most_frequency_K_words(T1)
 N=get_words_from_the_source_text(T2)
 C={} 
 for each word w in N:
  if w not in V or freq(w) <= threshold:
   I=get_synonyms_from_WordNet(w, $W_s$)
   for each i in I:
     if i not in V:
        I=I \ {i} #remove i from I
   sort_words_by_descend_of_frequency(I) 
   $C = C \cup \{w,I\}$
 return  C

n_best=3
apply_to_input_file(C, T2, n_best)
\end{lstlisting}
\label{LSW}
\end{minipage}
\section{Experiments}
We evaluate our approaches on the English-Vietnamese and the Japanese-Vietnamese translation systems. Translation performance is measured in BLEU~\cite{Papineni2012} by the multi-BLEU scripts from Moses\footnote{\url{https://github.com/moses-smt/mosesdecoder/tree/master/scripts}}.

\subsection{Datasets}
We consider two low-resource language pairs: Japanese-Vietnamese and English-Vietnamese. For Japanese-Vietnamese, we use the TED data provided by WIT3~\cite{cettoloEtAl:EAMT2012} and compiled by~\cite{Ngo2018}.  The training set includes 106758 sentence pairs, the validation and test sets are \textit{dev2010} (568 pairs) and \textit{tst2010} (1220 pairs). For English$\rightarrow$Vietnamese, we use the dataset from IWSLT 2015 \cite{Cettolo2015} with around 133K sentence pairs for the training set, 1553 pairs in \textit{tst2012} as the validation and 1268 pairs in \textit{tst2013} as the test sets.

For \textbf{\textit{LSW}} algorithm,  we crawled pairs of synonymous words from Japanese-English WordNet\footnote{\url{http://compling.hss.ntu.edu.sg/wnja/}} and achieved 315850 pairs for English and 1419948 pairs for Japanese.
\subsection{Preprocessing}
For English and Vietnamese, we tokenized the texts and then true-cased the tokenized texts using Moses script. We do not use any word segmentation tool for Vietnamese. For comparison purpose, Sennrich's BPE algorithm is applied for English texts. Following the same preprocessing steps for Japanese (\textit{JPBPE}) in \cite{Ngo2018}, we use KyTea\footnote{\url{http://www.phontron.com/kytea/}}~\cite{Neubig2011} to tokenize texts and then apply BPE on those texts. The number of BPE merging operators are 50k for both Japanese and English.

\subsection{Systems and Training} 
We implement our NMT systems using \textit{OpenNMT-py} framework\footnote{\url{https://github.com/OpenNMT/OpenNMT-py}} \cite{opennmt} with the same settings as in \cite{Ngo2018} for our baseline systems. Our system are built with two hidden layers in both encoder and decoder, each layer has 512 hidden units. In the encoder, a BiLSTM architecture is used for each layer and in the decoder, each layer are basically an LSTM layer. The size of embedding layers in both source and target sides is also 512. Adam optimizer is used with the initial learning rate of $0.001$ and then we apply learning rate annealing.  We train our systems for $16$ epochs with the batch size of 32. Other parameters are the same as the default settings of \textit{OpenNMT-py}.
\begin{center}
\begin{table*}[t]
\vspace*{-0.1cm}
{\small
\hfill{}
\begin{tabular}{|l|l|c|c|}
\hline
\hline
\multirow{2}{*}{\textbf{No.}} & \multirow{2}{*}{\textbf{Systems}} & \multicolumn{2}{c|}{\textbf{Japanese$\rightarrow$Vietnamese}}  \\
\cline{3-4}
 & & {\tt dev2010} & {\tt tst2010}  \\ 
 \hline
 (1) & Baseline  & 7.91 & 9.42   \\
 (2) & + Source Embedding & 7.77 & \textbf{9.96 }  \\
 (3) & + LSW & \textbf{8.37} & \textbf{10.34}   \\
 \hline
  (4) & JPBPE+VNBPE at Ngo et al (2018) & 7.77 & 9.04  \\
  (5) & JPBPE+VNBPE + BT + Mixsource at Ngo et al (2018) & 8.56 & 9.64   \\
 \hline
 \hline
 \multirow{2}{*}{\textbf{No.}} & \multirow{2}{*} {\textbf{Systems}} & \multicolumn{2}{c|}{\textbf{Vietnamese$\rightarrow$Japanese}}  \\
\cline{3-4}
 & & {\tt dev2010} & {\tt tst2010} \\ 
 \hline
 (1) & Baseline & 9.53 (9.53) & 10.95 (10.99) \\
 (2) & + Source Embedding & \textbf{10.51 (10.51)} & \textbf{11.37 (11.39)}   \\
 \hline
 (3) & JPBPE+VNBPE at Ngo et al (2018) & 9.74 & 11.13  \\
 \hline
 \hline
\end{tabular}}
\hfill{}
\caption{\label{resultsTED} {Results of Japanese-Vietnamese NMT systems}}

\end{table*}
\end{center}
We then modify the baseline architecture with the alternative proposed in Section~\ref{source-emb} in comparison to our baseline systems. All settings are the same as the baseline systems.
\subsection{Results}
In this section, we show the effectiveness of our methods on two low-resource language pairs and compare them to the other works. The empirical results are shown in Table~\ref{resultsTED} for Japanese-Vietnamese  and  in Table~\ref{resultsIWSLT} for  English-Vietnamese. Note that, the Multi-BLEU is only measured in the Japanese$\rightarrow$Vietnamese direction and the standard BLEU points are written in brackets.

\subsubsection{Japanese-Vietnamese Translation}
We conduct two out of the three proposed approaches for Japanese-Vietnamese translation systems and the results are given in the Table \ref{resultsTED}.

\textbf{Baseline Systems}.~~We find that our translation systems which use Sennrich's BPE method for Japanese texts and do not use word segmentation for Vietnamese texts are neither better or insignificant differences compare to those systems used word segmentation in \cite{Ngo2018}. Particularly, we obtained +0.38  BLEU points between (1) and (4) in the Japanese$\rightarrow$Vietnamese and -0.18 BLEU points between (1) and (3) in the Vietnamese$\rightarrow$Japanese.

\textbf{Our Approaches}.~On the systems trained with the modified architecture mentioned in the section~\ref{source-emb}, we obtained an improvements of +0.54 BLEU points in the Japanese$\rightarrow$Vietnamese and +0.42 BLEU points on the Vietnamese$\rightarrow$Japanese compared to the baseline systems. 

Due to the fact that Vietnamese WordNet is not available, we only exploit WordNet to tackle unknown words of Japanese texts in our Japanese$\rightarrow$Vietnamese translation system.  After using Kytea, Japanese texts are applied \textit{\textbf{LSW}} algorithm to replace OOV words by their synonyms. We choose 1-best synonym for each OOV word. Table \ref{TB-Number-JA} shows the number of OOV words replaced by their synonyms. The replaced texts are then BPEd and trained on the proposed architecture.  The largest improvement is +0.92 between (1) and (3). We observed an improvement of +0.7 BLEU points between (3) and (5) without using data augmentation described in~\cite{Ngo2018}.
\vspace{0.4cm}
\begin{table} [h]
\vspace*{-0.1cm}
\centerline{
\begin{tabular}{|c|c|c|c|}
\hline 
 & Train  &  dev2010 & tst2010  \\
\hline
 Number of words & 1015 & 36 & 25 \\
 \hline
\end{tabular}}
\caption{\label{TB-Number-JA} {The number of Japanese OOV words replaced by their synonyms.}}
\vspace*{-0.4cm}
\end{table}
\vspace{0.2cm}
\subsubsection{English-Vietnamese Translation}
We examine the effect of all approaches presented in Section~\ref{rare-word-translation} for our English-Vietnamese translation systems. Table \ref{resultsIWSLT} summarizes those results and the scores from other systems~\cite{Toan2017,Huang2018}.

\textbf{Baseline systems}. After preprocessing data using Moses scripts, we train the systems of English$\leftrightarrow$Vietnamese on our baseline architecture. Our translation system obtained +0.82 BLEU points compared to \cite{Toan2017} in the English$\rightarrow$Vietnamese and this is lower than the system of \cite{Huang2018} with neural phrase-based translation architecture.

\textbf{Our approaches}. The datasets from the baseline systems are trained on our modified NMT architecture. The improvements can be found as +0.55 BLEU points  between (1) and (2) in the English$\rightarrow$Vietnamese and +0.45 BLEU points (in \textit{tst2012}) between (1) and (2) in the Vietnamese$\rightarrow$English.
\begin{center}
\begin{table*}[ht]
\vspace*{-0.1cm}
{\small
\hfill{}
\begin{tabular}{|l|l|c|c|}
\hline
\hline
\multirow{2}{*}{\textbf{No.}} & \multirow{2}{*}{\textbf{Systems}} &  \multicolumn{2}{c|}{\textbf{English$\rightarrow$Vietnamese}} \\
\cline{3-4}
 & & {\tt tst2012} & {\tt tst2013} \\ 
 \hline
 (1) & Baseline  &  26.91 (24.39) & 29.86 (27.52)  \\
 (2) & + Source Embedding &  \textbf{27.41 (24.92)} & \textbf{30.41 (28.05)}  \\
 (3) & + Sennrich's BPE &  \textbf{26.96 (24.46)} & \textbf{30.10 (27.84)}  \\
 \hline
 (4) & + SAA & \textbf{27.16 (24.67}) & \textbf{30.60 (28.34)}  \\
 (5) & + LSW &  \textbf{27.46 (24.99)} & \textbf{30.85 (28.54)} \\
 \hline
 (6) & Nguyen and Chiang (2017)  & - &  26.7\\
 (7) & Huang et al (2018)   & - & 28.07 \\
 \hline
 \hline
 \multirow{2}{*}{\textbf{No.}} & \multirow{2}{*} {\textbf{Systems}} & \multicolumn{2}{c|}{\textbf{Vietnamese$\rightarrow$English}} \\
\cline{3-4}
 & & {\tt tst2012} & {\tt tst2013} \\ 
 \hline
 (1) & Baseline & 27.97 (28.52) & 30.07 (29.89) \\
 (2) & + Source Embedding & \textbf{28.42 (29.04)}& \textbf{30.12 (29.93)} \\
 \hline
 \hline
\end{tabular}}
\hfill{}
\caption{\label{resultsIWSLT} {Results  of  English-Vietnamese NMT systems}}
\end{table*}
\end{center}
\vspace{-0.58cm}
For comparison purpose, English texts are split into sub-words using Sennrich's BPE methods. We observe that, the achieved BLEU points are lower 
Therefore, we then apply the \textbf{\textit{SAA}} algorithm on the English texts from (2) in the English$\rightarrow$Vietnamese. The number of applied words are listed in Table \ref{TB-Number-SAA}. The improvement in BLEU are +0.74 between (4) and (1).
\vspace{0.2cm}
\begin{table} [h]
\vspace*{-0.1cm}
\centerline{
\begin{tabular}{|c|c|c|c|}
\hline 
 & Train  & tst2012  & tst2013  \\
\hline
 Number of words & 5342 & 84 & 93 \\
 \hline
\end{tabular}}
\caption{\label{TB-Number-SAA} {The number of rare words in which their affixes are detached from the English  texts in the SAA algorithm.}}
\vspace*{-0.2cm}
\end{table}

Similarly to the Japanese$\rightarrow$Vietnamese system, we apply \textbf{\textit{LSW}} algorithm on the English texts from (4) while selecting 1-best synonym for each OOV word. The number of replaced words on English texts are indicated in the Table~\ref{TB-Number-EN}. Again, we obtained a bigger gain of +0.99 (+1.02) BLEU points in English$\rightarrow$Vietnamese direction. Compared to the most recent work~\cite{Huang2018}, our system reports an improvement of +0.47 standard BLEU points on the same dataset.
\begin{table} [h]
\begin{center}
\begin{tabular}{|c|c|c|c|}
\hline 
 & Train  &  tst2012 & tst2013  \\
\hline
 Number of words & 1889 & 37 & 41 \\
 \hline
\end{tabular}
\caption{\label{TB-Number-EN} {The number of English OOV words are replaced by their synonyms.}}
\end{center}
\vspace*{-0.3cm}
\end{table}

We investigate some examples of translations generated by the English$\rightarrow$Vietnamese systems with our proposed methods in the Table \ref{TB-Exam}. The bold texts in red color present correct or approximate translations while the italic texts in gray color denote incorrect translations. The first example, we consider two words: \textit{presentation} and the unknown word \textit{applauded}. The word \textit{presentation} is predicted correctly as \foreignlanguage{Vietnamese}{\textit{"bài thuyết trình"}} in most cases when we combined source context through embeddings. The unknown word \textit{applauded} which has not seen in the vocabulary is ignored in the first two cases (baseline and source embedding) but it is roughly translated as \foreignlanguage{Vietnamese}{\textit{"hoan nghênh"}} in the \textit{\textbf{SAA}} because it is separated into \textit{applaud} and \textit{ed}. In the second example, we observe the translations of the unknown word \textit{tryout}, they are mistaken in the first three cases but in the \textit{\textbf{LSW}}, it is predicted with a closer meaning as  \foreignlanguage{Vietnamese}{"bài kiểm tra"} due to the replacement by its synonymous word as  \textit{test}. 

\begin{center}
\begin{table*}[ht]
\vspace*{-0.1cm}
{\small
\hfill{}
\begin{tabular}{p{3cm} p{12.5cm}}
\hline 
 Source   &   ~~which {\color{red}\textbf{presentation}} have you {\color{red}\textbf{applauded}} the most this morning ? \\
 Reference &  \foreignlanguage{Vietnamese}{~~{\color{red}\textbf{bài thuyết trình}} nào bạn {\color{red}\textbf{vỗ tay}} nhiều nhất trong sáng nay ? } \\
 Baseline &  \foreignlanguage{Vietnamese}{~ {\color{gray}\textit{điều này}} có thể diễn ra trong buổi sáng hôm nay ? } \\
 +Source Embedding & \foreignlanguage{Vietnamese}{~~{\color{red}\textbf{bài thuyết trình}} nào có thể tạo ra buổi sáng hôm nay ?} \\
 +SAA &  \foreignlanguage{Vietnamese}{~ {\color{red}\textbf{bài thuyết trình}} này có {\color{red}\textbf{hoan nghênh}} buổi sáng hôm nay không ?} \\
 +LSW & \foreignlanguage{Vietnamese}{~~{\color{gray}\textit{điều gì}} đã diễn ra với bạn buổi sáng hôm nay ?} \\
 \hline
  Source   &  ~~I started this as a {\color{red} \textbf{tryout}} in Esperance , in Western Australia . \\
 Reference &  \foreignlanguage{Vietnamese}{~~tôi đã bắt đầu như một {\color{red} \textbf{sự thử nghiệm} } tại Esperance , tây Úc . } \\
 Baseline &  \foreignlanguage{Vietnamese}{~~tôi bắt đầu như thế này như là một {\color{gray}\textit{người đàn ông}} , ở phương Tây Úc . }\\
 +Source Embedding & \foreignlanguage{Vietnamese}{~~ tôi đã bắt đầu điều này như là một {\color{gray}\textit{người đàn áp}} ở ven biển ở Tây Úc . } \\
 +SAA &  \foreignlanguage{Vietnamese}{~~tôi đã bắt đầu như thế này với tư cách là một {\color{gray}\textit{người đàn ông}} trong lĩnh vực này, ở Tây Úc . }\\
 +LSW & \foreignlanguage{Vietnamese}{~~tôi bắt đầu thí nghiệm này như một {\color{red} \textbf{bài kiểm tra}} ở Quảng trường , ở Tây Úc .} \\
 \hline
\end{tabular}}
\hfill{}
\caption{\label{TB-Exam} {Examples of outputs from the English$\rightarrow$Vietnamese translation systems with the proposed methods.}}
\end{table*}
\end{center}
\vspace{-0.6cm}
\section{Related Works}
Addressing unknown words was mentioned early in the Statistical Machine Translation (SMT) systems. Some typical studies as: \cite{NizarHabash2008} proposed four techniques to overcome this situation by extend the morphology and spelling of words or using a bilingual dictionary or transliterating for names. These approaches are difficult when manipulate to different domains. \cite{TrieuLong2016} trained word embedding models to learn word similarity from monolingual data and an unknown word are then replaced by a its  similar word. \cite{Madhyastha2017} used a linear model to learn maps between source and target spaces base on a small initial bilingual dictionary to find the translations of source words. However, in NMT, there are not so many works tackling this problem. \cite{Jean2015} use a very large vocabulary to solve unknown words. \cite{Luong2015b} generate a dictionary from alignment data based on annotated corpus to decide the hypotheses of unknown words. 
\cite{Toan2017} have introduced the solutions for dealing with the rare word problem, however, their models require more parameters, thus, decreasing the overall efficiency.

In another direction, \cite{Sennrich2016} exploited the BPE algorithm to reduce number of unknown words in NMT and achieved significant efficiency on many language pairs. The second approach presented in this works follows this direction when instead of using an unsupervised method to split rare words and unknown words into sub-words that are able to translate, we use a supervised method. Our third approach using WordNet can be seen as a smoothing way, when we use the translations of the synonymous words to approximate the translation of an OOV word. Another work followed this direction is worth to mention is~\cite{niehues2016using}, when they use the morphological and semantic information as the factors of the words to help translating rare words.
\section{Conclusion}
In this study, we have proposed three difference strategies to handle rare words in NMT, in which the combination of methods brings significant improvements to the NMT systems on two low-resource language pairs. In future works, we will consider selecting some appropriate synonymous words for the source sentence from n-best synonymous words to further improve the performance of the NMT systems and leverage more unsupervised methods based on monolingual data to address rare word problem.
\section{Acknowledgments}
This work is supported by the project "Building a machine translation system to support translation of documents between Vietnamese and Japanese to help managers and businesses in Hanoi approach to Japanese market", No. TC.02-2016-03.
\bibliography{emnlp-ijcnlp-2019}
\bibliographystyle{acl_natbib}

\end{document}